\documentclass[twoside,11pt]{article}

\usepackage{jmlr2e,url}
\usepackage{amsmath,amssymb}
\usepackage{graphics,color}

\usepackage[hide]{simple_notes}

\newcommand{\BEAS}{\begin{eqnarray*}}
\newcommand{\EEAS}{\end{eqnarray*}}
\newcommand{\BEA}{\begin{eqnarray}}
\newcommand{\EEA}{\end{eqnarray}}
\newcommand{\BEQ}{\begin{equation}}
\newcommand{\EEQ}{\end{equation}}
\newcommand{\BIT}{\begin{itemize}}
\newcommand{\EIT}{\end{itemize}}
\newcommand{\BNUM}{\begin{enumerate}}
\newcommand{\ENUM}{\end{enumerate}}
\newcommand{\BA}{\begin{array}}
\newcommand{\EA}{\end{array}}
\newcommand{\diag}{\mathop{\rm diag}}
\newcommand{\Diag}{\mathop{\rm Diag}}

\newcommand{\rank}{\mathop{\rm rank}}

\newcommand{\tr}{\mathop{ \rm tr}}

\newcommand{\rb}{\mathbb{R}}

\newcommand{\mysec}[1]{Section~\ref{sec:#1}}
\newcommand{\eq}[1]{Eq.~(\ref{eq:#1})}
\newcommand{\myfig}[1]{Figure~\ref{fig:#1}}

\newcommand {\br}[1]{\left(#1\right)}

\newcommand {\cbr}[1]{\left\{#1 \right\}}
\newcommand {\nm}[1]{\Arrowvert\, #1 \,\Arrowvert}
\newcommand {\inp}[1]{\left\langle #1 \right\rangle}
\newcommand {\inpH}[2]{\left\langle #1 \right\rangle_{#2}}

\newcommand {\ovr}[1]{\frac{1}{#1}}

\newcommand{\RR}{\mathbb{R}}    
\newcommand{\NN}{\mathbb{N}}    
\newcommand{\Gcal}{\mathcal{G}}
\newcommand{\Hcal}{\mathcal{H}}

\newcommand{\Xcal}{\mathcal{X}} 
\newcommand{\Ycal}{\mathcal{Y}} 
\newcommand{\Bcal}{\mathcal{B}} 
\newcommand{\Vcal}{\mathcal{V}} 
 
\newcommand{\xb}{\mathbf{x}}    
\newcommand{\yb}{\mathbf{y}}    
\newcommand{\hb}{\mathbf{h}}    
\newcommand{\ub}{\mathbf{u}}    
\newcommand{\vb}{\mathbf{v}}    

\newcommand{\kD}{k_{\rm D  } }    
\newcommand{\kA}{k_{\rm A  } }    

\newcommand{\OMIT}[1]{}

\title{A New Approach to Collaborative Filtering: \\
Operator Estimation with Spectral Regularization}

\author{\name Jacob Abernethy \email jake@cs.berkeley.edu \\
 \addr
Division of Computer Science\\
University of California \\
387 Soda Hall, Berkeley, CA, USA \\
 \AND
  \name Francis R. Bach \email francis.bach@mines.org\\
\addr
INRIA - WILLOW Project-Team \\
Laboratoire d'Informatique de l'Ecole Normale Sup\'erieure
(CNRS/ENS/INRIA UMR 8548) \\
45, rue d'Ulm,
75230 Paris, France \\
\AND
\name
Theodoros Evgeniou \email theodoros.evgeniou@insead.edu \\
\addr
Decision Sciences and Technology Management \\
INSEAD \\ 
Bd de Constance, 77300 Fontainebleau, France\\
\AND
\name
Jean-Philippe Vert \email Jean-Philippe.Vert@Mines-ParisTech.fr \\
\addr
Centre for Computational Biology\\
Mines ParisTech, Institut Curie, Inserm U900\\
35 rue Saint-Honor{\'e}, 77300 Fontainebleau, France\\
}

\date{\today}

\begin{document}

\maketitle

 \begin{abstract}

We present a general approach for collaborative filtering (CF) using
spectral regularization to learn linear operators from ``users'' to
a set of possibly desired ``objects''. Recent low-rank type matrix completion
approaches to CF are shown to be special cases. However, unlike
existing regularization based CF methods, our approach can be used to
also incorporate information such as attributes of the users or the objects---a limitation of existing regularization based CF methods. We
provide novel representer theorems that we use to develop new
estimation methods. We then provide learning algorithms based on low-rank
decompositions, and test them on a standard CF dataset.  The
experiments indicate the advantages of generalizing the existing
regularization based CF methods to incorporate related information
about users and objects. Finally, we show that certain multi-task
learning methods can be also seen as special cases of our proposed
approach.

\end{abstract}

\section{Introduction}

Collaborative filtering (CF) refers to the task of predicting
preferences of a given ``user'' for some ``objects'' (e.g., books,
music, products, people, etc.) based on his/her previously revealed
preferences---typically in the form of purchases or ratings---as well as the
revealed preferences of other users. In a book recommender system, for
example, one would like to suggest new books to someone based on what
she and other users have recently purchased or rated. The ultimate goal
of CF is to infer the preferences of users in order to offer them new objects. 

A number of CF methods have been developed in the past
\citep{CF1,CF2,CF3}. Recently
there has been interest in CF using regularization based methods
\citep{Srebro2003Weighted}. This work adds to that literature by developing a novel
general approach to developing regularization based CF methods.

Recent regularization based CF methods assume that the only data available are
the revealed preferences, where no other information such as background
information on the objects or users is given. In this case, one may formulate
the problem as that of inferring the contents of a partially observed
\emph{preference matrix}: each row represents a user, each column represents
an object (e.g., books or movies), and entries in the matrix represent a given
user's rating of a given object. When the only information available is a set
of observed user/object ratings, the unknown entries in the matrix must be
inferred from the known ones -- of which there are typically very few relative
to the size of the matrix.

To make useful predictions within this setting, regularization based
CF methods make certain assumptions about the {\em relatedness} of the
objects and users. The most common assumption is that preferences can
be decomposed into a small number of factors, both for users and
objects, resulting in the search for a low-rank matrix which
approximates the partially observed matrix of preferences
\citep{Srebro2003Weighted}. The rank constraint can be interpreted as
a regularization on the hypothesis space. Since the rank constraint
gives rise to a non-convex set of matrices, the associated
optimization problem will be a difficult non-convex problem for which
only heuristic algorithms exist \citep{Srebro2003Weighted}. An
alternative formulation, proposed by~\citet{Srebro2005Maximum},
suggests penalizing the predicted matrix by its {\em trace norm},
i.e., the sum of its singular values. An added benefit of the
trace norm regularization is that, with a sufficiently large
regularization parameter, the final solution will be
low-rank~\citep{Boyd2001Rank,bach-tracenorm}.

However, a key limitation of current regularization based CF methods
is that they do not take advantage of information, such as attributes
of users (e.g., gender, age) or objects (e.g., book's author, genre),
which is often available. Intuitively, such information might be
useful to guide the inference of preferences, in particular for users
and objects with very few known ratings. For example, at the extreme,
users and objects with no prior ratings can not be considered in the
standard CF formulation, while their attributes alone could provide
some basic preference inference.

The main contribution of this paper is to develop a general framework and
specific algorithms also based on novel representer theorems for the more general CF setting where other information,
such as attributes for users and/or objects, may be available. More precisely
we show that CF, while typically seen as a problem of matrix completion, can
be thought of more generally as estimating a linear operator from the space of
users to the space of objects. Equivalently, this can be viewed as learning a
bilinear form between users and objects. We then develop {\it spectral
regularization} based methods to learn such linear operators. When dealing
with operators, rather than matrices, one may also work with infinite
dimension, allowing one to consider arbitrary feature space, possibly induced by some
kernel function. Among key theoretical contributions of this paper are new
representer theorems, allowing us to develop new general methods that learn finitely
many parameters even when working in infinite dimensional 
user/object feature space. These representer theorems generalize
the classical representer theorem for minimization of an empirical loss
penalized by the norm in a Reproducing Kernel Hilbert Space (RKHS) to more
general penalty functions and function classes.

We also show that, with the appropriate choice of kernels for both users and
objects, we may consider a number of existing machine learning methods as
special cases of our general framework. In particular, we show that several CF
methods such as rank constrained optimization, trace-norm regularization, and those based on
Frobenius norm regularization, can all be cast as special cases of spectral
regularization on operator spaces. Moreover, particular choices of kernels
lead to specific sub-cases such as regular matrix completion and multitask
learning. In the specific application of collaborative filtering with the
presence of attributes, we show that our generalization of these sub-cases
leads to better predictive performance.

The outline of the paper is as follows. In Section~\ref{sec:notation}, we
review the notion of a compact operator on Hilbert Space, and we show how to
cast the collaborative filtering problem within this framework. We then
introduce spectral regularization and discuss how rank constraint, trace norm
regularization, and Frobenius norm regularization are all special cases of
spectral regularization. In Section~\ref{sec:related}, we show how our general
framework encompasses many existing methods by proper choices of the loss
function, the kernels, and the spectral regularizer. In
Section~\ref{sec:repthms}, we provide three representer theorems for operator
estimation with spectral regularization which allow for efficient learning
algorithms. Finally in Section~\ref{sec:algs} we present a number of
algorithms and describe several techniques to improve efficiency. We test these
algorithms in Section~\ref{sec:exp} on synthetic examples and a widely used
movie database.

\section{Learning compact operators with spectral regularization}\label{sec:notation}

In this section we propose a mathematical formulation for a general CF
problem with spectral regularization. We then show in \mysec{related} how several learning problems can be cast under this general framework.

\subsection{A general CF formulation}\label{eq:cf}
We consider a general CF problem in which our goal is to model the preference of a user described by $\xb$ for an item described by $\yb$. We denote by $\xb$ and $\yb$ the data objects containing all relevant or available information; this could, for example, include a unique identifier $i$ for the $i$-th user or object. Of course, the users and objects may additionally be  characterized by attributes, in which case $\xb$ or $\yb$ would contain some representation of this extra information. Ultimately, we would like to consider such attribute information as encoded in some positive definite kernel between users, or equivalently between objects. This naturally leads us to model the users as elements in a Hilbert space $\Xcal$, and the objects they rate as elements of another Hilbert space $\Ycal$. 

We assume that our observation data is in the form of {\it ratings} from users to objects, a real-valued score representing the user's preference for the object. Alternatively, similar methods can be applied when the observations are binary, specifying for instance whether or not a user considered or selected an object.

Given a series of $N$ observations
$(\xb_{i},\yb_{i},t_{i})_{i=1,\ldots,N}$ in
$\Xcal\times\Ycal\times\RR$, where $t_{i}$ represents the rating of
user $\xb_{i}$ for object $\yb_{i}$, the generalized CF
problem is then to infer a function $f:\Xcal\times\Ycal \rightarrow
\RR$ that can then be used to predict the rating of any user
$\xb\in\Xcal$ for any object $\yb\in\Ycal$ by $f\br{\xb,\yb}$. Note
that in our notation, $\xb_{i}$ and $\yb_{i}$ represent the
user and object corresponding to the $i$-th rating available. If several ratings of a user for different objects
are available, as is commonly the case, several $\xb_{i}$'s will be identical in $\Xcal$---a slight abuse of notation. 
We denote by $\Xcal_{N}$ and $\Ycal_{N}$ the linear spans
of $\cbr{\xb_{i}\,,\,i=1,\ldots,N}$ and
$\cbr{\yb_{i}\,,\,i=1,\ldots,N}$ in $\Xcal$ and $\Ycal$, with
respective dimensions $m_{\Xcal}$ and $m_{\Ycal}$.

For the function to be estimated we restrict ourselves to bilinear forms given by:
\begin{equation}\label{eq:bilinear}
f(\xb,\yb) = \inpH{\xb,F\yb}{\Xcal}\,,
\end{equation}
for some compact operator $F$. We now denote by $\Bcal_{0}\br{\Ycal,\Xcal}$ the set of compact operators from $\Xcal$ to $\Ycal$. For an introduction to relevant concepts in functional analysis, see Appendix~\ref{app:functional}.

In the general case we consider below, if $\Xcal$ and $\Ycal$ are not Hilbert spaces, one could
also first map (implicitly) users $\xb$ and objects $\yb$ into possibly infinite
dimensional Hilbert feature spaces $\Phi_{\Xcal}(\xb)$ and $\Psi_{\Ycal}(\yb)$
and use kernels. We
refer the reader to Appendix A for basic definitions and properties
related to compact operators that are useful below. The inference
problem can now be stated as follows:

\emph{Given a training set of ratings, how may we estimate a ``good'' compact operator $F$ to predict future ratings using (\ref{eq:bilinear})?}

 We estimate the operator $F$ in (\ref{eq:bilinear}) from the training
data using a standard regularization and statistical machine learning
approach. In particular, we propose to define the operator as the
solution of an optimization problem over $\Bcal_{0}\br{\Ycal,\Xcal}$
whose objective function balances a data fitting term $R_{N}(F)$,
which is small for operators that can correctly explain the training
data, with a regularization term $\Omega(F)$. We now describe these
two terms in more details.

\subsection{Data fitting term}
Given a loss function $\ell(t',t)$ that quantifies how good a
prediction $t' \in\RR$ is if the true value is $t\in\RR$, we consider a
fitting term equal to the empirical risk, i.e., the mean loss incurred
on the training set:
\begin{equation}\label{eq:empiricalrisk}
R_{N}(F) = \ovr{N}\sum_{i=1}^N \ell\br{\inp{\xb_{i},F\yb_{i}}_
\Xcal, t_{i}}\,.
\end{equation}
The particular choice of the loss function should typically depend on
the precise problem to be solved and on the nature of the variables
$t$ to be predicted. See more details in \mysec{related}.
 In particular, while the representer theorems presented in \mysec{repthms} do not need any convexity with respect to this choice, the algorithms presented in \mysec{algs} do.

\subsection{Regularization term}
\label{sec:spectralpenalties}
For the regularization term, we focus on a class of spectral functions defined as follows.
\begin{definition}
A function $\Omega : \Bcal_{0}\br{\Ycal,\Xcal} \mapsto \RR \cup
\cbr{+\infty}$ is called a \emph{spectral penalty function} if it can
be written as:
\begin{equation}\label{eq:spectralloss}
\Omega(F) = \sum_{i=1}^d s_{i}\br{\sigma_{i}(F)}\,,
\end{equation}
where for any $i\geq 1, s_{i}:\RR^+\mapsto\RR^+\cup\cbr{+\infty}$ is a
non-decreasing penalty function satisfying $s(0)=0$, and $\left(\sigma_i(F)\right)_{i=1,\ldots,d}$ are the $d$ singular values of $F$ in decreasing order---$d$ possibly infinite.
\end{definition}
Note that by the spectral theorem presented in Appendix~\ref{app:functional}, any compact operator can be decomposed into  singular vectors, with singular values being a sequence that tends to zero.

Spectral penalty functions include as special cases several functions
often encountered in matrix completion problems:
\begin{itemize}
\item For a given integer $r$, taking $s_{i}=0$ for $i=1,\ldots,r$ and $s_{r+1}(u)=+\infty$ if $u>0$, leads to the function:
\begin{equation}\label{eq:rankpenalty}
\Omega(F) = 
\begin{cases}
0 & \text{ if }\rank(F)\leq r\,,\\
+\infty & \text{ otherwise.}
\end{cases}
\end{equation}
 
In other words, the set of operators $F$ that satisfy $\Omega(F)<+\infty$ is the set of operators with rank smaller than $r$.
\item Taking $s_{i}(u) = u$ for all $i$ results in the trace norm penalty (see Appendix~\ref{app:functional}):
\begin{equation}\label{eq:tracepenalty}
\Omega(F) = 
\begin{cases}
\nm{F}_{1} & \text{ if }F \in \Bcal_{1}\br{\Ycal,\Xcal},\\
+\infty & \text{ otherwise,}
\end{cases}
\end{equation}
where we note with $\Bcal_{1}\br{\Ycal,\Xcal}$ the set of operators with finite trace norm. Such operators are referred to as trace class operators.

\item Taking $s_{i}(u) = u^2$ for all $i$ results in the squared Hilbert-Schmidt norm penalty (also called squared Frobenius norm for matrices, see Appendix~\ref{app:functional}):
\begin{equation}\label{eq:hspenalty}
\Omega(F) = 
\begin{cases}
\nm{F}_{2}^2 & \text{ if }F \in \Bcal_{2}\br{\Ycal,\Xcal},\\
+\infty & \text{ otherwise,}
\end{cases}
\end{equation}
where we note with $\Bcal_{2}\br{\Ycal,\Xcal}$ the set of operators with finite squared Hilbert-Schmidt norm. Such operators are referred to as  Hilbert Schmidt operators.

\end{itemize}

These particular functions can be combined together in different ways. For example, we may constrain the rank to be smaller than $r$ while penalizing the trace norm of the matrix, which can be obtained by setting $s_{i}(u) = u$ for $i=1,\ldots,r$ and $s_{r+1}(u)=+\infty$ if $u>0$. Alternatively, if we want to penalize the Frobenius norm while constraining the rank, we set $s_{i}(u) = u^2$ for $i=1,\ldots,r$ and $s_{r+1}(u)=+\infty$ if $u>0$. We state these two choices of $\Omega$ explicitly since we use these in the experiments~(see \mysec{exp}) or to design efficient algorithms~(see \mysec{algs}):

\begin{eqnarray}\label{eq:tracerankpenalty}
\text{Trace+Rank Penalty:} & &
\Omega(F) = 
\begin{cases}
\nm{F}_{1} & \text{ if }\rank(F)\leq r,\\
+\infty & \text{ otherwise.}
\end{cases}\\
\label{eq:frobrankpenalty}
\text{Frobenius+Rank Penalty:} & &
\Omega(F) = 
\begin{cases}
\nm{F}_{2}^2 & \text{ if }\rank(F)\leq r,\\
+\infty & \text{ otherwise.}
\end{cases}
\end{eqnarray}

\subsection{Operator inference}
With both a fitting term and a regularization term, we can now
formally define our inference approach. It consists of finding an
operator $\hat{F}$, if there exists one, that solves the following
optimization problem:
\begin{equation}\label{eq:theproblem}
\hat{F} \in \underset{F \in \Bcal_{0}\br{\Ycal,\Xcal}}{\arg\min}   R_{N}(F) + \lambda \Omega(F)\,,
\end{equation}
where $\lambda\in\RR$ is a parameter that controls the trade-off
between fitting and regularization, and where $R_{N}(F)$ and
$\Omega(F)$ are respectively defined in (\ref{eq:empiricalrisk}) and
(\ref{eq:spectralloss}). We note that if the set
$\cbr{F \in \Bcal_{0}\br{\Ycal,\Xcal} \, , \,\Omega(F)<+\infty}$ is not empty, then necessarily the
solution $\hat{F}$ of this optimization problem must satisfy
$\Omega(\hat{F})<\infty$.

We show in Sections \ref{sec:repthms} and \ref{sec:algs} how
problem (\ref{eq:theproblem}) can be solved in practice in particular 
for Hilbert spaces of  infinite dimensions. Before exploring such 
implementation-related issues, in the following section 
we provide several examples of algorithms that can be
derived as particular cases of (\ref{eq:theproblem}) and highlight
their relationships to existing methods.

\section{Examples and related approaches} 
\label{sec:related}

The general formulation (\ref{eq:theproblem}) can result in a variety
of practical algorithms potentially useful in different contexts.  In
particular, three elements can be tailored to one's particular needs:
the loss function, the kernels (or equivalently the Hilbert spaces), and the spectral penalty term. We
start this section by some generalities about the possible choices for
these elements and their consequences, before highlighting some
particular combinations of choices relevant for different
applications.
\begin{enumerate}
\item {\bf The loss function.} The choice of $\ell$ defines the
empirical risk through (\ref{eq:empiricalrisk}). It is a classical
component of many machine learning methods, and should typically
depend on the type of data to be predicted (e.g., discrete or
continuous) and of the final objective of the algorithm (e.g.,
classification, regression or ranking). The choice of $\ell$ also
influences the algorithm, as discussed in Section \ref{sec:algs}. As a
deeper discussion about the loss function is only tangential to the
current work, we only consider the square loss here, knowing that other convex losses may be considered.  

\item {\bf The spectral penalty function.} The choice of $\Omega(F)$
defines the type of constraint we impose on the operator that we seek
to learn. In Section \ref{sec:spectralpenalties}, we gave several
examples of such constraints including the rank constraint
(\ref{eq:rankpenalty}), the trace norm constraint
(\ref{eq:tracepenalty}), the Hilbert-Schmidt norm constraint
(\ref{eq:hspenalty}), or the trace norm constraint over low-rank
operators (\ref{eq:tracerankpenalty}). The choice of a particular
penalty might be guided by some considerations about the problem to be
solved, e.g., finding low-rank operators as a way to discover
low-dimensional latent structures in the data. On the other hand, from
an algorithmic perspective, the choice of the spectral penalty may
affect the efficiency or feasibility of our learning
algorithm. Certain penalty functions, such as the rank constraint for
example, will lead to non-convex problems because the corresponding
penalty function (\ref{eq:rankpenalty}) is not convex itself. However, the same rank constraint can vastly reduce the number
of parameters to be learned. These algorithmic considerations are
discussed in more details in Section \ref{sec:algs}.

\item {\bf The kernels.} Our choice of kernels defines the inner
products (i.e., embeddings) of the users and objects in their
respective Hilbert spaces. We may use a variety of possible kernels depending on the problem to be solved and on the attributes
available. Interestingly, the choice of a particular kernel has no
influence on the algorithm, as we show later (however, it does of course influence the running time of these algorithms).  In the current work, we focus on two basic kernels (Dirac kernels and attribute kernels) and in Section \ref{sec:corners} we discuss combining these.

\begin{itemize}
\item The first kernel we consider is the \emph{Dirac} kernel. When two
users (resp. two objects) are compared, the Dirac kernel returns
$1$ if they are the same user (resp. object), and $0$
otherwise. In other words, the Dirac kernel amounts to representing the
users (resp. the objects) by orthonormal vectors in $\Xcal$
(resp.~in $\Ycal$). This kernel can be used whether or not attributes
are available for users and objects. We denote by
$k_{D}^{\Xcal}$ (resp.~$k_{D}^{\Ycal}$) the Dirac kernel for the
users (resp.~objects).

\item The second kernel we consider is a kernel between
\emph{attributes}, when attributes are available to describe the
users and/or objects. We call this an ``attribute kernel''.  This
would typically be a kernel between vectors, such as the inner product
or a Gaussian RBF kernel, when the descriptions of users and/or
objects take the form of vectors of real-valued attributes, or any kernel on structured objects~\citep{Cristianini2004}. We denote
by $k_{A}^{\Xcal}$ (resp. $k_{A}^{\Ycal}$) the attributes kernel for
the users (resp. objects).
\end{itemize}
\end{enumerate}

In the following section we illustrate how specific combinations of loss, spectral
penalty and kernels can be relevant for various settings. In particular the choice
of kernels leads to new methods for a range of different estimation problems;
namely, matrix completion, multi-task learning, and pairwise learning. In
Section~\ref{sec:corners} we consider a new representation that allows
interpolation between these particular problem formulations.

\subsection{Matrix completion}
When the Dirac kernel is used for both users and objects, then we can  organize the data $\{ \xb_i, i =
1,\dots,n\}$ into $n_\Xcal$ groups of identical data points and
similarly $\{ \yb_i, i = 1,\dots,n\}$ into $n_\Ycal$ groups.
Since we use the Dirac kernel, we can represent each of these groups by the elements
of the canonical basis
$\br{\ub_{1},\ldots,\ub_{n_{\Xcal}}}$ and
$\br{\vb_{1},\ldots,\vb_{n_{\Ycal}}}$   of
$\RR^{n_{\Xcal}}$ and $\RR^{n_{\Ycal}}$, respectively. A bilinear form using Dirac kernels only depends on the identities of the users and the objects, and we only predict the rating $t_i$ based on the
identities of the groups in both spaces.
 If we assume that each pair user/object is observed at most once, the data can be
re-arranged into a $n_\Xcal \times n_\Ycal$ incomplete matrix, the
learning objective being to complete this matrix (indeed, in this
context, it is not possible to generalize to never seen points in
$\Xcal$ and $\Ycal$). 

In this case, our bilinear form framework exactly corresponds to completing the matrix, since
the bilinear function of $\xb$ and $\yb$ is exactly equal to
$\ub_{i}^\top M \vb_{j}$ where $\xb = \ub_{i}$ (i.e., $\xb$ is the $i$-th person) and
$\yb = \vb_{j}$ (i.e., $\yb$ is the $j$-th object). Thus,  the $(i,j)$-th entry of
the matrix $M$ can be assimilated to the value of the bilinear form
defined by the matrix $M$ over the pair $(\ub_{i},\vb_{j})$. Moreover
 the spectral regularizer corresponds to the corresponding spectral function of the complete matrix $M \in \rb^{ n_\Xcal 
\times n_\Ycal}$.

 In this context, finding a low-rank
approximation of the observed entries in a matrix is an appealing
strategy, which corresponds to taking the rank penalty constraint
(\ref{eq:rankpenalty}) combined with, for example, the square loss
error. This however leads to non-convex optimization problems with
multiple local minima, for which only local search heuristics are
known \citep{Srebro2003Weighted}. To circumvent this issue, convex
spectral penalty functions can be considered. Indeed, in the case
of binary preferences, combining the hinge loss function with the
trace norm penalty (\ref{eq:tracepenalty}) leads to the maximum margin
matrix factorization (MMMF) approach proposed by
\citet{Srebro2005Maximum}, which can be rewritten as a semi-definite
program. For the sake of efficiency, \citet{Rennie2005Fast} proposed
to add a constraint on the rank of the matrix, resulting in a
non-convex problem that can nevertheless be handled efficiently by
classical gradient descent techniques; in our setting, this
corresponds to changing the trace norm penalty (\ref{eq:tracepenalty})
by the penalty (\ref{eq:tracerankpenalty}).

\subsection{Multi-task learning}

It may be the case that we have attributes only for objects $\yb$ (we could do the same for attributes for users). In that case, for a finite number of users
$\cbr{\xb_{i}\, , \,i=1,\ldots,N}$ organized in $n_\Xcal$ groups, we aim to estimate a separate function on objects
$f_i(\yb)$ for each of the $n_\Xcal$ users $i$. Considering the estimation of each of these
$f_i$'s as a \emph{learning task}, one can possibly learn all $f_i$'s
\emph{simultaneously} using a \emph{multi-task learning} approach.

In order to adapt our general framework to this scenario, it is
natural to consider the attribute kernel $k_{A}^\Ycal$ for the
objects, whose attributes are available, and the Dirac kernel
$k_{D}^\Xcal$ for the users, for which no attributes are
used. Again the choice of the loss function depends on the precise
task to be solved, and the spectral penalty function can be tuned to
enforce some sharing of information between different tasks.

In particular, taking the rank penalty function (\ref{eq:rankpenalty})
enforces a decomposition of the tasks (learning each $f_i$) into a
limited number of factors. This results in a method for multitask
learning based on a low-rank representation of the predictor
functions $f_i$. The resulting problem, however, is not convex due to
the use of the non-convex rank penalty function. A natural alternative
is then to replace the rank constraint by the trace norm penalty
function~(\ref{eq:tracepenalty}), resulting in a convex optimization
problem when the loss function is convex. Recently, a similar approach
was independently proposed by \citet{srebro-mc} in the context of
multiclass classification and by \citet{Argyriou2008Convex} for
multitask learning.

Alternatively, another strategy to enforce some constraints among the
tasks is to constrain the variance of the different
classifiers. \citet{Evgeniou2005Learning} showed that this strategy
can be formulated in the framework of support vector machines by
considering a \emph{multitask kernel}, i.e., a kernel $k_{multitask}$ over the product space $\Xcal\times\Ycal$ defined between any two user/object pairs $\br{\xb,\yb}$ and $\br{\xb',\yb'}$ by:
\begin{equation}\label{eq:mtkernel}
k_{multitask}\br{\br{\xb,\yb},\br{\xb,\yb}} = \br{k_{D}^\Xcal\br{\xb,\xb'}+c}k_{A}^{\Ycal}\br{\yb,\yb'}\,,
\end{equation}
where $c>0$ controls how the variance of the classifiers is constrained compared to the norm of each classifier. As explained in Appendix A, estimating a function over the product space $\Xcal\times\Ycal$ by penalizing the RKHS norm of the kernel (\ref{eq:mtkernel}) is a particular case of our general framework, where we take the Hilbert-Schmidt norm (\ref{eq:hspenalty}) as spectral penalty function, and where the kernels between users and between objects are respectively $k_{D}^\Xcal\br{\xb,\xb'}+c$ and $k_{A}^{\Ycal}\br{\yb,\yb'}$. When $c=0$, i.e., when we take a Dirac kernel for the users and an attribute kernel for the objects, then penalizing the Hilbert-Schmitt norm amounts estimating independent models for each users, as explained in \citet{Evgeniou2005Learning}. Combining two Dirac kernels for users and objects, respectively, and penalizing the Hilbert-Schmitt norm would not be very interesting, since the solution would always be $0$ except on the training pairs. On the other hand, replacing
the Hilbert-Schmidt norm defined by other penalties such as the trace norm
penalty \eqref{eq:tracepenalty} would be an interesting extension when the kernels $k_{D}^\Xcal\br{\xb,\xb'}+c$ and $k_{A}^{\Ycal}\br{\yb,\yb'}$ are used: this would constrain both the variance of the
predictor functions $f_i$ and their decomposition into a small number of factors, which could be an interesting approach in some multitask learning applications.

\subsection{Pairwise learning}
When attributes are available for both users and objects then
it is possible to take the attributes kernels for both of
them. Combining this choice with the Hilbert-Schmidt penalty
\eqref{eq:hspenalty} results in classical machine learning algorithms
(e.g., an SVM if the hinge loss is taken as the loss function) applied to
the \emph{tensor product} of $\Xcal$ and $\Ycal$. This strategy is a
classical approach to learn a function over pairs of points \citep[see, e.g.,][]{Jacob2006Epitope}. Replacing
the Hilbert-Schmidt norm by another spectral penalty function, such as
the trace norm, would result in new algorithms for learning low-rank
functions over pairs.

\subsection{Combining the attribute and Dirac kernels}\label{sec:corners}
As illustrated in the previous subsections, the setting of the
application often determines the combination of kernels to be used for
the users and the objects: typically, two Dirac kernels for the
standard CF setting without attributes, one Dirac and one attributes
kernel for multi-task problems, and two attributes kernels when
attributes are available for both users and objects and one
wishes to learn over pairs.

There are many situations, however, where the attributes available to
describe the users and/or objects are certainly useful for the
inference task, but on the other hand do not fully characterize the
users and/or objects. For example, if we just know the age and gender
of users, we would like to use this information to model their
preferences, but would also like to allow different preferences for
different users even when they share the same age and gender. In our
setting, this means that we may want to use the attributes kernel in
order to utilize known attributes from the users and objects during inference, but also
the Dirac kernel to incorporate the fact that different users and/or objects
remain different even when they share many or all of their attributes.

This naturally leads us to consider the following convex combinations
of Dirac and attributes kernels~\citep{lowrank}: 
\begin{equation}\label{eq:kernelcombination}
\begin{cases}
k^\mathcal{X} = \eta \kA^\Xcal + (1-\eta) \kD^\Xcal ,\\
k^\Ycal = \zeta \kA^\Ycal  + (1-\zeta) \kD^\Ycal ,\\
\end{cases}
\end{equation}
where $0\leq\eta\leq 1$ and $0\leq\zeta\leq 1$. These kernels
interpolate between the Dirac kernels ($\eta=0$ and $\zeta=0$) and the
attributes kernels ($\eta=1$ and $\zeta=1$). Combining this choice of
kernels with, e.g., the trace norm penalty function
(\ref{eq:tracepenalty}), allows us to continuously interpolate between
different settings corresponding to different ``corners'' in the
$(\eta,\zeta)$ square: standard CF with matrix completion in $(0,0)$,
multi-task learning in $(0,1)$ and $(1,0)$, and learning over pairs in
$(1,1)$. The extra degree of freedom created when $\eta$ and $\zeta$
are allowed to vary continuously between $0$ and $1$ provides a
principled way to optimally balance the influence of the attributes in
the function estimation process.

 \note{FB: the following paragraph is mostly here to answer the reviewers' comments. It mainly says that what we do is essentially blocking some of the columns of the low-rank decomsosition but slightly better because we have direct control on the tradeoffs.}
 
Note that our representational framework encompasses  simpler natural approaches to include attribute information for collaborative filtering: for example, one could consider completing matrices using matrices of the form 
$UV^\top + U_A R_A^\top + U_A S_A^\top$, where $UV^\top$ is a low-rank matrix to be optimized, $U_A$ and $V_A$ are the given attributes for the first and second domains, and $R_A$, $S_A$ are parameters to be learned. This formulation corresponds to adding an unconstrained low-rank term $UV^\top$, and  the simpler linear predictor from the concatenation of attributes  $U_A R_A^\top + U_A S_A^\top$~\citep{Jacob2006Epitope}. Our approach implicitly adds a fourth cross-product term $U_A T V_A^\top$, where $T$ is estimated from data.
This exactly corresponds to imposing that the low rank matrix has a decomposition which includes $U_A$ and $V_A$ as columns. Our combination of Dirac and attribute kernels has the advantage of having specific weights $\eta$ and $\zeta$ that control the trade-off between the constrained and unconstrained low-rank matrices.

%
%

\section{Representer theorems} \label{sec:repthms}
We now present the key theoretical results of this paper and discuss how the general optimization problem (\ref{eq:theproblem}) can be solved in practice. A first difficulty with this problem is that the optimization space $\cbr{ F \in\Bcal_{0}\br{\Ycal,\Xcal}\,:\,\Omega(F)<\infty}$ can be of infinite dimension. We note that this can occur even under a rank constraint, because the set $\cbr{ F \in\Bcal_{0}\br{\Ycal,\Xcal}\,:\, \rank(F)\leq R}$ is not included into any finite-dimensional linear subspace if $\Xcal$ and $\Ycal$ have infinite dimensions. In this section, we show that the optimization problem (\ref{eq:theproblem}) can be rephrased as a finite-dimensional problem, and propose practical algorithms to solve it in Section \ref{sec:algs}. While the reformulation of the problem as a finite-dimensional problem is a simple instance of the representer theorem when the Hilbert-Schmidt norm is used as a penalty function (Section \ref{sec:hsrepresenter}), we prove in Section \ref{sec:generalrepresenter} a generalized representer theorem that is valid with any spectral penalty function.

\subsection{The case of the Hilbert-Schmidt penalty function}\label{sec:hsrepresenter}

In the particular case where the penalty function $\Omega(F)$ is the Hilbert-Schmidt norm (\ref{eq:hspenalty}), then the set $\cbr{F \in\Bcal_{0}\br{\Ycal,\Xcal}\,:\,\Omega(F)<\infty}$ is the set of Hilbert-Schmidt operators. As recalled in Appendix~\ref{app:functional}. this set is a Hilbert space isometric through (\ref{eq:bilinear}) to the reproducing kernel Hilbert space $\Hcal_{\otimes}$ of the kernel:
$$
k_{\otimes}\br{\br{\xb,\xb'},\br{\yb,\yb'}} = \inpH{\xb,\xb'}{\Xcal}\inpH{\yb,\yb'}{\Ycal}\,,
$$
and the isometry translates from $F$ to $f$ as:
$$
\nm{f}_{\Hcal_{\otimes}}^2 = \nm{F}^2 = \Omega(F)\,.
$$
As a result, in that case the problem (\ref{eq:theproblem}) is equivalent to:
\begin{equation}\label{eq:theproblem2}
\underset{f\in\Hcal_{\otimes}}{\min} \cbr{ R_{N}(f) + \lambda \nm{f}_{\otimes}^2}\,.
\end{equation}
In that case the representer theorem for optimization of empirical risks penalized by the RKHS norm \citep{Aronszajn1950Theory,representer} can be applied to show that the solution of (\ref{eq:theproblem2}) necessarily lives in the linear span of the training data. With our notations this translates into the following result:
\begin{theorem}\label{th:hilbertschmidt}
If $\hat{F}$ is a solution of the problem:
\begin{equation}
\underset{F \in \Bcal_{2}\br{\Ycal,\Xcal} }{\min}   R_{N}(F) + \lambda \sum_{i=1}^\infty \sigma_{i}(F)^2 \,,
\end{equation}
then it is necessarily in the linear span of $\cbr{\xb_{i}\otimes \yb_{i}\,:\,i=1,\ldots,N}$, i.e., it can be written as:
\begin{equation}\label{eq:expansion_classical}
\hat{F} = \sum_{i=1}^N \alpha_{i} \xb_{i}\otimes \yb_{i}\,,
\end{equation}
for some $\alpha\in\RR^N$.
\end{theorem}
For the sake of completeness, and to highlight why this result is specific to the Hilbert-Schmidt penalty function (\ref{eq:hspenalty}), we rephrase here, with our notations, the main arguments in the proof of \citet{representer}. Any operator $F$ in $\Bcal_{2}\br{\Ycal,\Xcal}$ can be decomposed as $F = F_{S} + F_{\perp}$, where $F_{S}$ is the projection of $F$ onto the linear span of $\cbr{\xb_{i}\otimes \yb_{i}\,:\,i=1,\ldots,N}$. $F_{\perp}$ being orthogonal to each $\xb_{i}\otimes\yb_{i}$ in the training set, one easily gets $R_{N}(F) = R_{N}(F_{S})$, while $\nm{F}^2 = \nm{F_{S}}^2 + \nm{F_{\perp}}^2$ by the Pythagorean theorem. As a result a minimizer $F$ of the objective function must be such that $F_{\perp}=0$, i.e., must be in the linear span of the training tensor products.

\subsection{A Representer Theorem for General Spectral Penalty Functions}\label{sec:generalrepresenter}

Let us now move on to the more general situation (\ref{eq:theproblem}) where a general spectral function $\Omega(F)$ is used as regularization. Theorem \ref{th:hilbertschmidt} is usually not valid in such a case. Its proof breaks down because it is not true that $\Omega(F) = \Omega(F_{S}) + \Omega(F_{\perp})$ for general $\Omega$, or even that $\Omega(F)\geq\Omega(F_{S})$.

The following theorem, whose proof is presented in Appendix~\ref{app:proof}, can be seen as a generalized representer theorem. It shows that a solution of (\ref{eq:theproblem}), if it exists, can be expanded over a finite basis of dimension $m_{\Xcal}\times m_{\Ycal}$ (where $m_\Xcal$ and $m_\Ycal$ are the underlying dimensions of the subspaces where the data lie), and that it can be found as the solution of a finite-dimensional optimization problem (with no convexity assumptions on the loss):

\begin{theorem}\label{th:representerspectral}
For any spectral penalty function $\Omega:\Bcal_{0}\br{\Ycal,\Xcal} \mapsto \RR\cup\cbr{+\infty}$, let the optimization problem:
\begin{equation}\label{eq:optspectral}
\underset{F \in \Bcal_{0}\br{\Ycal,\Xcal} ,}{\min} R_{N}(F) + \lambda \Omega(F) \,.
\end{equation}
If the set of solutions is not empty, then there is a solution $F$ in $\Xcal_{N}\otimes \Ycal_{N}$, i.e., there exists $\alpha\in\RR^{m_{\Xcal}\times m_{\Ycal}}$ such that:
\begin{equation}\label{eq:expansion}
F = \sum_{i=1}^{m_{\Xcal}} \sum_{j=1}^{m_{\Ycal}} \alpha_{ij} \ub_{i}\otimes \vb_{j}\,,
\end{equation}
where $\br{\ub_{1},\ldots,\ub_{m_{\Xcal}}}$ and $\br{\vb_{1},\ldots,\vb_{m_{\Ycal}}}$ form orthonormal bases of $\Xcal_{N}$ and $\Ycal_{N}$, respectively. Moreover, in that case the coefficients $\alpha$ can be found by solving the following finite-dimensional optimization problem:
\begin{equation}\label{eq:optalpha}
\min_{\alpha\in\RR^{m_{\Xcal}\times m_{\Ycal}} }
R_{N}\br{\diag\br{X \alpha Y^\top}} + \lambda \Omega(\alpha)\,,
\end{equation}
where $\Omega(\alpha)$ refers to the spectral penalty function applied to the matrix $\alpha$ seen as an operator from $\RR^{m_{\Ycal}}$ to $\RR^{m_{\Xcal}}$, and $X \in \RR^{N\times m_{\Xcal}}$ and $Y\in \RR^{N\times m_{\Xcal}}$ denote any matrices that satisfy $K = XX^\top$ and $G = YY^\top$ for the two $N \times N$ Gram matrices $K$ and $G$ defined by $K_{ij}= \inpH{\xb_{i},\xb_{j}}{\Xcal}$ and $G_{ij}= \inpH{\yb_{i},\yb_{j}}{\Ycal}$, for $0\leq i,j\leq N$.
\end{theorem}

This theorem shows that, as soon as a spectral penalty function is used to control the complexity of the compact operators, a solution can be searched in the finite-dimensional space $\Xcal_{N}\otimes \Ycal_{N}$, which in practice boils down to an optimization problem over the set of matrices of size $m_{\Xcal}\times m_{\Ycal}$. The dimension of this space might however be prohibitively large for real-world applications where, e.g., tens of thousands of users are confronted to a database of thousands of objects. A convenient way to obtain an important decrease in complexity (at the expense of possibly losing convexity) is by constraining the rank of the operator through an adequate choice of a spectral penalty. Indeed, the set of non-zero singular components of $F$ as an operator is equal to the set of non-zero singular values of $\alpha$ in (\ref{eq:expansion}) seen as a matrix. Consequently any constraint on the rank of $F$ as an operator results in a constraint on 
 $\alpha$ as a matrix, from which we deduce: 
\begin{corollary}
If, in Theorem \ref{th:representerspectral}, the spectral penalty function $\Omega$ is infinite on operators of rank larger than $R$ (i.e., $\sigma_{R+1}(u)=+\infty$ for $u>0$), then the matrix $\alpha\in\RR^{m_{\Xcal}\times m_{\Ycal}}$ in (\ref{eq:expansion}) has rank at most $R$.
\end{corollary}

As a result, if a rank constraint $\rank(F)\leq r$ is added to the optimization problem then the representer theorem still holds but the dimension of the parameter $\alpha$ becomes $r\times\br{m_{\Xcal}+m_{\Ycal}}$ instead of $m_{\Xcal}\times m_{\Ycal}$, which is usually beneficial. We note, however, that when a rank constraint is added to the Hilbert-Schmidt norm penalty, then the classical representer Theorem \ref{th:hilbertschmidt} and the expansion of the solution over $N$ vectors (\ref{eq:expansion_classical}) are not valid anymore, only Theorem \ref{th:representerspectral} and the expansion (\ref{eq:expansion}) can be used.

\section{Algorithms}
 \label{sec:algs}

In this section we explain how the optimization problem (\ref{eq:optalpha}) can be solved in practice. We first consider a general formulation, then we specialize to the situation
where many $\xb$'s and many $\yb$'s are identical; i.e., we are in a matrix completion
setting where it may be advantageous to consider other formulations that take into account some group structure explicitly.

\subsection{Convex dual of spectral regularization}
When the loss is convex, we can derive the convex dual problem, which can be helpful for actually solving the optimization problem. This could also provide an alternative proof of the representer theorem in that particular situation.

For all $i=1,\dots,N$, we let denote
$\psi_i(v_i) = \ell\br{ v_i , t_i } $ the loss corresponding to predicting $v_i$ for the $i$-th data point. For simplicity, we
assume that each $\psi_i$ is convex (this is usually met in practice). Following~\citet{bach_thibaux}, we let $\psi_i^\ast(\alpha_i)$ denote its Fenchel conjugate defined
as $\psi_i^\ast(\alpha_i) = \max_{v_i \in \RR} \alpha_i v_i - \psi_i(v_i)$. Minimizers of the optimization problem defining the conjugate function are often referred to as Fenchel duals to $\alpha_i$~\citep{Boyd2003Convex}.
In particular, we have the following classical examples:
\BIT
\item \emph{Least-squares regression}: we have
 $\psi_i(v_i)= \frac{1}{2}(t_i - v_i)^2$ and $\psi_i^\ast(\alpha_i) = \frac{1}{2} \alpha_i^2 + \alpha_i t_i$.
 \item \emph{Logistic regression}: we have
 $\psi_i(v_i)=\log(1+\exp(-y_i v_i))$, where $y_i \in \{-1,1\}$,
 and $\psi_i^\ast(\alpha_i) = (1+\alpha_i t_i) \log(1+\alpha_i t_i) - \alpha_i t_i \log(-\alpha_i t_i)$
if $\alpha_i t_i \in (-1,0)$, $+ \infty$ otherwise.
\EIT

We also assume 
that the spectral regularization is such that for all $i \in \NN$, $s_i=s$, where $s$ is a 
convex function such that $s(0)=0$. In this situation, we have
$\Omega(A) = \sum_{i \in \NN} s(\sigma_i(A))$. We can also define a Fenchel conjugate for $\Omega(A)$, which is also a spectral function $\Omega^\ast(B) = \sum_{i \in \NN
} s^\ast(\sigma_i(B))$~\citep{lewis02twice}.

Some special cases of interest for $s(\sigma)$ are: 
\BIT
\item $s(\sigma) = |\sigma|$ leads to the trace norm and
then $s^\ast(\tau) = 0$ if $|\tau|$ is less than 1, and $+\infty$ otherwise.  

\item $s(\sigma) = \frac{1}{2} \sigma^2$ leads to the Frobenius/Hilbert Schmidt norm and
then $s^\ast(\tau) = \frac{1}{2} \tau^2$.  

\item $s(\sigma) = \varepsilon \log ( 1+ e^{\sigma/\varepsilon} ) + \varepsilon \log ( 1+ e^{-\sigma/\varepsilon} )$  
is a smooth approximation of $|\sigma|$, which becomes tighter when $\varepsilon$ is closer to zero. We have:
$s^\ast(\tau) = \frac{1}{\varepsilon} ( 1 + \tau ) \log(1+\tau) +
\frac{1}{\varepsilon} ( 1 - \tau ) \log(1-\tau)$. Moreover,
$s'(\sigma) = \tau \Leftrightarrow  (s^\ast)'(\tau) = \sigma = \frac{1}{\varepsilon}
\log \frac{ 1+\tau}{1-\tau}
$.

\EIT

Once the representer theorem has been applied, our optimization problem
can be rewritten in the \emph{primal} form in 
(\ref{eq:optalpha}):
\BEQ
\label{eq:primal_spectral}
\min_{\alpha \in \RR^{m_x 
\times m_y}} 
\sum_{i=1}^N \psi_i( (X\alpha Y^\top)_{ii} )  + \lambda \Omega(\alpha).
\EEQ
We can now form the Lagrangian, associated with added constraints $v = \diag( X\alpha Y^\top)$ and corresponding Lagrange multiplier $\beta \in \rb^N$:
$$ \mathcal{L}(v,\alpha,\beta)
= \sum_{i=1}^N \psi_i( v_i )  - \sum_{i=1}^N\beta_i ( v_i -
(X\alpha Y^\top)_{ii} ) + \lambda \Omega(\alpha),
$$
and minimize with respect to $v$ and $W$ to obtain the \emph{dual} problem, which is to maximize:
\BEQ
\label{eq:dual_spectral}
 - \sum_{i=1}^N \psi^\ast_i( \beta_i ) - \lambda  \Omega^\ast \left( - \frac{1}{\lambda} X^\top \Diag(\beta) Y  \right).
\EEQ
Once the optimal dual variable $\beta$ is found (there are as many of those as there are observations),
then we can go back to $\alpha$ (which may or may not be of smaller size), by Fenchel duality, i.e., $\alpha$ is among the Fenchel duals of $
 - \frac{1}{\lambda} X^\top \Diag(\beta) Y$. Thus, when the function $s$ is differentiable and strictly convex (which implies that the set of Fenchel duals is a singleton), then we obtain the primal variables  $\alpha$ in closed form from the dual variables $\beta$. When $s$ is not differentiable, e.g., for the trace norm then, following~\citet{srebro-mc}, we can find the primal variables by noting that once $\beta$ is known, the singular vectors of $\alpha$ are known and we can find the singular values by solving a reduced convex optimization problem.

     \paragraph{Computational complexity} 
 Note that for optimization, we have two strategies: using the primal problem in \eq{primal_spectral} of dimension $m_\Xcal m_\Ycal  \leqslant n_\Xcal n_\Ycal  $ (the actual dimension of the underlying  data) or using the dual problem
 in \eq{dual_spectral} of dimension $N$ (the number of ratings). The choice between those two formulations is problem dependent.

  \subsection{Collaborative filtering}
 In the presence of (many) identical columns and rows, which is often the case in collaborative filtering situations, the kernel matrices $K$ and $L$ have some columns (and thus rows) which are identical,
 and we can instead consider the kernel matrices (with their square-root decompositions) $\tilde{K} = \tilde{X} \tilde{X}^\top$ and $\tilde{L} = \tilde{Y} \tilde{Y}^\top$
 as the kernel matrices for all distinct elements of $\mathcal{X}$ and $\mathcal{Y}$ (let 
 $n_\mathcal{X}$ and $n_\mathcal{Y}$ be their sizes). Then each observation $(\xb_i,\yb_i,t_i)$ corresponds to a pair of indices $(a(i),b(i))$ in $\{1,\dots,n_\mathcal{X}\}
 \times \{1,\dots,n_\mathcal{Y}\}$, and the primal/dual problems become:

\begin{equation} \label{eq:primal}
\min_{\alpha \in \RR^{m_x 
\times m_y}} 
\sum_{i=1}^n \psi_i( \delta_{a(i)}^\top \tilde{X} \alpha \tilde{Y}^\top\delta_{b(i)} )  + \lambda \Omega(\alpha),
\end{equation}
where $\delta_{u}$ is a vector with only zeroes except at position $u$. The dual function is
$$ - \sum_{i=1}^N \psi^\ast_i( \beta_i ) - \lambda  \Omega^\ast \left( - \frac{1}{\lambda} \tilde{X}^\top \sum_{i=1}^N \beta_i \delta_{a(i)}\delta_{b(i)}^\top \tilde{Y}  \right).
$$
Similar to usual kernel machines and the general case presented above, using the primal or the dual formulation for optimization depends on the number of available ratings $N$ compared to the ranks $m_\mathcal{X}$ and $m_\mathcal{Y}$ of the kernel matrices $\tilde{K}$ and $\tilde{L}$. Indeed, the number of variables in the primal formulation is $m_\mathcal{X} m_\mathcal{Y}$, while in the dual formulation it is $N$.

\subsection{Low-rank constrained problem}
\label{sec:lowrank}
We approximate the spectral norm by an infinitely differentiable spectral function. Since we consider in this paper only infinitely differentiable loss functions, our problem is that of minimizing an infinitely differentiable convex function $G(W)$ over rectangular matrices
of size $p \times q$ for certain integers $p$ and $q$. As a result of our spectral regularization, we hope to obtain (approximately) low-rank matrices. In this context, it has proved advantageous to 
consider low-rank decompositions of the form $W = UV^\top$ where $U$ and $V$ have $m<\min \{p,q\}$ columns~\citep{burer1,burer2}. \citet{burer1} have shown that if $m=\min\{p,q\}$ then the non-convex problem of minimizing $G(UV^\top)$ with respect
to $U$ and $V^\top$ has no local minima.

We now prove a stronger result in the context of twice differentiable functions, namely that if the global optimum of $G$ has rank $r<\min \{p,q\}$, then the low-rank constrained
problem with rank $r+1$ has no local minimum and its global minimum  corresponds to the
global minimum of $G$. The following theorem makes this precise (see Appendix~\ref{app:lowrank} for proof).

\begin{proposition}
\label{prop:conv_lowrank}
Let $G$ be a   twice differentiable convex function on matrices of
size $p \times q$ with compact level sets. Let $m>1$ and
$(U,V) \in \rb^{p  \times m} \times \rb^{q  \times m}  $ a local optimum of the function $H: \rb^{p  \times m} \times \rb^{q  \times m}  \mapsto \rb$ defined
by $H(U,V)=G(UV^\top)$, i.e.,  $U$ is such that $\nabla H (U,V) = 0 $ and the Hessian of $H$ at $(U,V)$ is positive semi-definite. If $U$ or $V$ is rank deficient, then $N=UV^\top$ is a global  minimum of $G$, i.e., $\nabla G(N)  =0$.
\end{proposition}

The previous proposition shows that if we have a local minimum for the rank-$m$ problem and if the solution is
rank deficient, then we have a solution of the global optimization problem. This naturally leads to   a sequence of reduced problems
of increasing dimension $m$,    smaller than $r+1$, where $r$ is the rank of the global optimum. However, the number of iterations of each of the local minimizations and
the final rank $m$ cannot be bounded a priori in general.

Note that using a low-rank representation to solve the trace-norm regularized problem leads to a non-convex minimization problem with no local minima, while simply using the low-rank representation \emph{without} the trace norm penalty and potentially with a Frobenius norm penalty, may lead to local minima; i.e., we consider instead of 
\eq{optalpha} with the trace norm, the following formulation:

\begin{equation}\label{eq:optalpha_lowrank_frobenius}
\min_{\alpha\in\RR^{m_{\Xcal} \times r}, \
\beta \in \RR^{   m_{\Ycal} \times r} }
R_{N}\br{ \diag\br{X \alpha \beta^\top Y^\top}} + \lambda  \sum_{q=1}^r \| \alpha(:,k) \|^2
 \| \beta(:,k) \|^2 \,,
\end{equation}
where $ \alpha(:,k) $ and $ \beta(:,k) $ are the $k$-th columns of $\alpha$ and $\beta$.
In the simulation section, we compare the two approaches on a synthetic example, and show that the convex formulation solved through a sequence of non-convex formulations leads to better predictive performance.

\subsection{Kernel learning for spectral functions}
In our collaborative filtering context, there are two potentially useful sources of kernel learning: learning the attribute kernels, or learning the weights $\eta$ and $\zeta$ between Dirac kernels and attribute kernels.  In this section, we show how multiple kernel learning (MKL)~\citep{gert,BachLanc2004skm} may be extended to spectral regularization.

We first show that the optimization problem that we have defined in earlier sections only depends on the Kronecker product of kernel matrices $K \otimes G$:

\begin{proposition}
The dual solution of the optimization problem in \eq{dual_learning} depends only on the matrix $K \otimes G$.
\end{proposition}
\begin{proof}
It suffices to show that for all matrices $B$, then the positive singular values of $X^\top B Y $ only depend on $K\otimes G$. The largest singular value is defined as the maximum of
$a^\top X^\top B Y b$ over unit norm vectors $a$ and $b$. By a change of variable, it is equivalent to maximize $\frac{ (X^\top \tilde{a}) X^\top B Y  (Y^\top\tilde{b})}{\| X^\top\tilde{a}\| \| Y^\top\tilde{b} \|    } = 
\frac{   {\rm vec}( \tilde{b} \tilde{a}^\top) (K \otimes G) {\rm vec}(B)  }{ {\rm vec}( \tilde{b} \tilde{a}^\top)^\top (K \otimes G) {\rm vec}( \tilde{b} \tilde{a}^\top)   }
 $ with respect to $\tilde{a}$ and $\tilde{b}$~\citep{golub83matrix}. Thus the largest positive singular value is indeed a function of $K \otimes G$. Results for other singular values may be obtained similarly.
\end{proof}

This shows that the natural kernel matrix to be learned in our context is the Kronecker product $K \otimes G$. We thus follow \citet{gert} and consider $M$ kernel matrices $K_1,\dots, K_M$ for $\mathcal{X}$ and $M$
kernel matrices  $G_1,\dots, G_M$ for $\mathcal{Y}$; one possibility could be to learn a convex combination of the matrices $K_k \otimes G_k$ by minimizing with respect to the combination weights the optimal value of the problem in \eq{dual_learning}. However, unlike the usual Hilbert norm regularization, this does not lead to a convex problem in general. We thus focus on the alternative formulation of the MKL problem~\citep{BachLanc2004skm}: we consider the sum of the predictor functions associated with each of the individual kernel pairs $(K_k,G_k)$ and penalize by the sum of the norms.

That is, if we let denote 
 $X_1,\dots,X_M$ and  $Y_1,\dots,Y_M$ the respective square roots of matrices
 $K_1,\dots, K_M$ and   $G_1,\dots, G_M$, we look for predictor functions which  are  sums of the $M$ possible atomic predictor functions, and we  penalize
by the sum of spectral functions, to obtain the following optimization problem:
$$
\min_{\forall k, \alpha_k \in \RR^{m_x^k
\times m_y^k}} 
\sum_{i=1}^n \psi_i \left( \sum_{k=1}^M (X_k \alpha_k Y_k^\top)_{ii} \right)  + \lambda \sum_{k=1}^M\Omega(\alpha_k).
$$
We form the Lagrangian:
$$ \mathcal{L}(v,\alpha_1,\dots,\alpha_M,\beta)
= \sum_{i=1}^n \psi_i( v_i )  - \sum_{i=1}^N\beta_i ( v_i -
\sum_{k=1}^M (X\alpha_k Y^\top)_{ii} ) + \lambda \sum_{k=1}^M \Omega(\alpha_k),
$$
and minimize w.r.t. $v$ and $\alpha_1,\dots,\alpha_M$ to obtain the dual problem, which is to maximize
\BEQ
\label{eq:dual_learning}
 - \sum_u \psi^\ast_i( \beta_i  ) - \sum_k \lambda   \Omega^\ast \left( - \frac{1}{\lambda} 
 X_k^\top \Diag(\beta) Y_k    \right).
\EEQ
In the case of the trace norm, we obtain support kernels~\citep{BachLanc2004skm}, i.e., only a sparse combination of matrices ends up being used. Note that in the dual formulation,
there is only one $\alpha$ to optimize, and thus it is preferable to use the dual formulation
rather than the primal formulation. 

This framework can be naturally applied to combine the four corners defined in \mysec{corners}. Indeed, we can form $M=4$ kernel matrices for each of the four corners and learn a combination of such matrices. We show in \mysec{exp} how the MKL framework allows to automatically combine these four corners without setting the trade-off directly though $\eta$ and $\zeta$ (by the user or through cross-validation).

\section{Experiments} \label{sec:exp}

In this Section we present several experimental findings for the algorithms and methods discussed above. Much of the present work was motivated by the problem of collaborative filtering and we therefore focus solely within this domain. As discussed in Section~\ref{sec:related}, by using operator estimation and spectral regularization as a framework for CF, we may utilize potentially more information to predict preferences. Our primary goal now is to show that, as one would hope, such capabilities do improve prediction accuracy.

\subsection{Datasets and Metrics}
 
We present several plots created by experimenting on synthetic data. This dataset was generated
as follows: (1) sample i.i.d.~multivariate features for $x$ of dimension 6, (2) generate
i.i.d.~multivariate features for $y$ of dimension 6 as well, (3) sample $z$ from a random
bilinear form in $x$ and $y$ plus some noise, (4) restrict the observed feature space to only 3
features for both $x$ and $y$. Since part of the data is discarded, the label cannot be
perfectly predicted by the known features. On the other hand, since we keep some of them,
knowing and using these attributes should work better than not using them. In other words, we
expect that setting $\eta$ and $\zeta$ to be values other than 0 or 1 should provide better
performance. 

We also experimented with the well-known MovieLens 100k dataset from the GroupLens Research Group at the University of Minnesota. This dataset consists of ratings of 1682 movies by 943 users. Each user provided a rating, in the form of a score from $\{1,2,3,4,5\}$, for a small subset of the movies. Each user rated at least 20 movies, and the total number of ratings available is exactly 100,000, averaging about 105 per user. This dataset was rather appropriate as it included attribute information for both the movies and the users. Each movie was labeled with at least one among $19$ genres (e.g., action or adventure), while the users' attributes included age, gender, and an occupation among a list of $21$ occupations (e.g., administrator or artist). We converted the users' age attribute to a set of  binary features that describes to which of 5 age categories the user belongs.

All test set accuracies are measured as the root mean squared error averaged over
10-fold cross validations. In particular, we focus on the comparisons of intermediate values of
$\eta$ and $\zeta$, compared to the four ``corners'' of the $\eta/\zeta-$parameter space:
\begin{itemize} 
        \item $\eta=0, \zeta=0$: matrix completion 
        \item $\eta=0, \zeta=1$ and $\eta=1, \zeta=0$: multi-task learning on users or objects 
        \item $\eta=1, \zeta=1$: pairwise learning
\end{itemize}

\subsection{Results}

\paragraph{Tracenorm Versus Low-rank} \label{sec:traceVSrank}

\begin{figure}
\begin{center}
\includegraphics[width=2.7in]{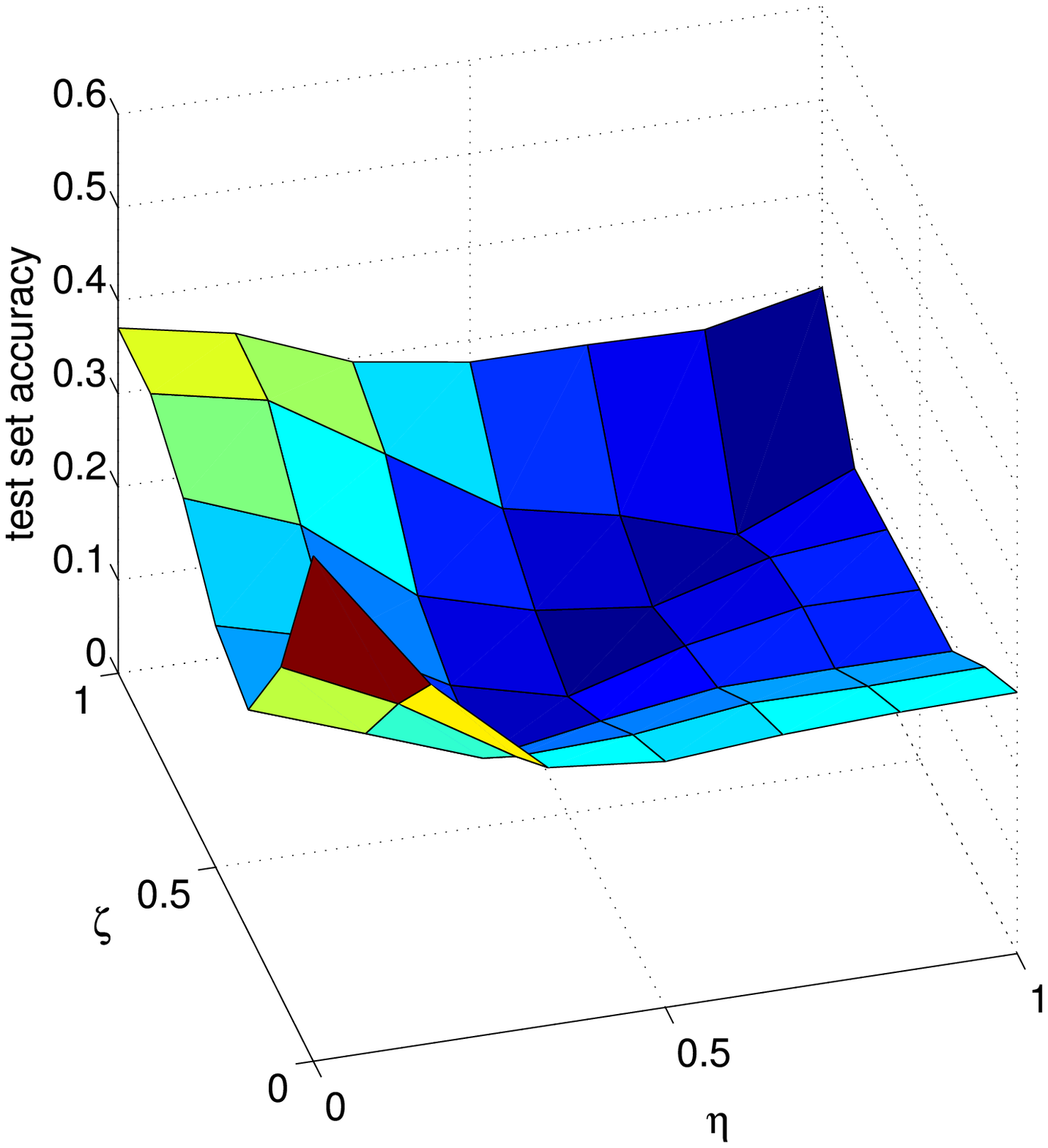}
\includegraphics[width=2.7in]{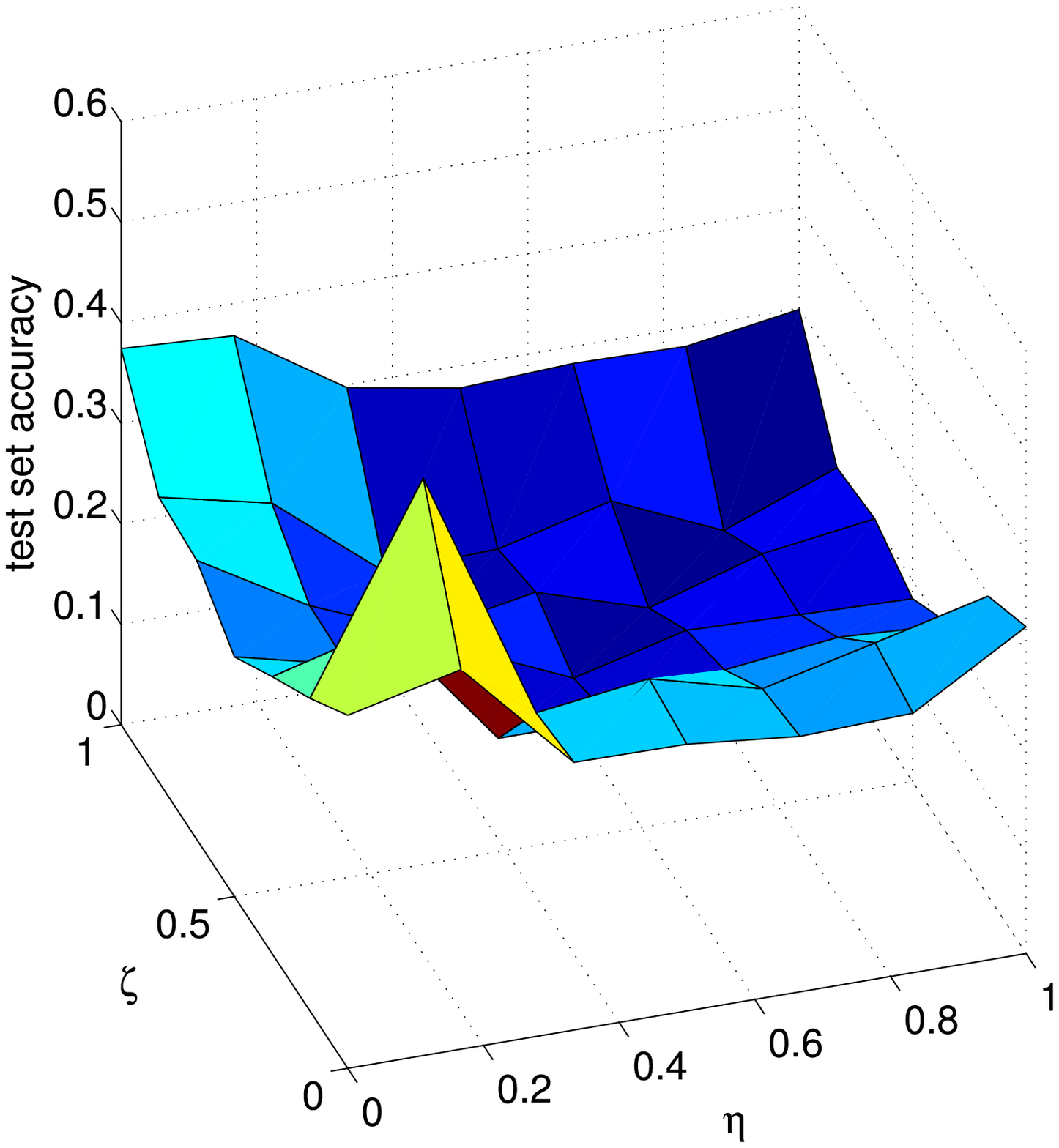}
\end{center}

\caption{Comparison between two spectral penalties: the trace norm (left) and the Frobenius
norm (right), each with an additional fixed rank constraint as described in
Section~\ref{sec:lowrank}. Each surface plot displays performance values over a range of $\eta$
and $\zeta$ values, all obtained using the synthetic dataset. The minimal value achieved by the
trace norm is 0.1222 and the one achieved by the rank constraint is 0.1540.}

\label{fig:toy_tracenorm_vs_rank}
\end{figure}

In \myfig{toy_tracenorm_vs_rank}, we present two performance plots over the $\eta/\zeta$
parameter space, both obtained using the synthetic dataset. The left plot displays the
results when utilizing the trace norm spectral penalty. Here we used the low rank decomposition
formulation described in Section~\ref{sec:lowrank} which (by
Proposition~\ref{prop:conv_lowrank}) has no local minima. The plot on the right utilizes the
same rank-constrained formulation, but with a Frobenius norm penalty instead. The
trace norm constrained algorithm performs slightly better.
Moreover, best predictive performance is achieved in both cases in the middle of the square and
not at any of the four corners.

 \note{FB: reviewer 2 makes the interesting point that it is not clear whether the fixed rank formulation is underperforming because (a) it cannot find the global optimum or (b) because this global minimum is in fact worse. He/she suggests to initialize the fixed-rank constraints with a projection of the trace norm solution. This requires to run some simulations. \inote{JA: It's a good point, but I don't think we want/need to run more experiments.} \inote{FB: OK, but it then requires some writing}}

\begin{figure}
\begin{center}
\includegraphics[scale=.7]{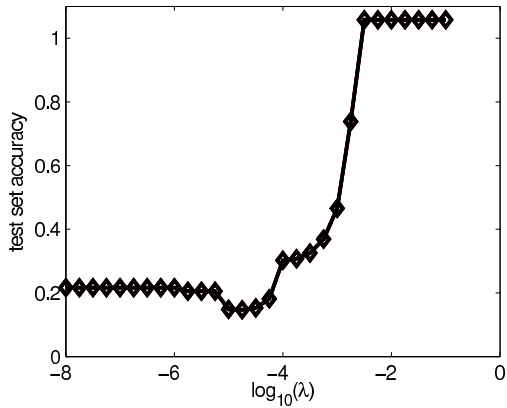}
\end{center}
\caption{Learning the kernel: test set accuracy vs. regularization parameter. Minimum value is 0.14.}
\label{fig:mkl}
\end{figure}

\paragraph{Kernel Learning} 
In \myfig{mkl}, we show the test set accuracy as a function of the regularization parameter, when we use the kernels corresponding to the four corners as the four basis kernels. We can see that we recover similar performance (error of 0.14 instead of 0.12) than by searching over all $\eta$ and $\zeta$'s. The same algorithm could also be used to learn kernels on the attributes.

\paragraph{Performance on MovieLens Data}
 
\begin{figure} 
        \begin{center}
                \includegraphics[width=4in]{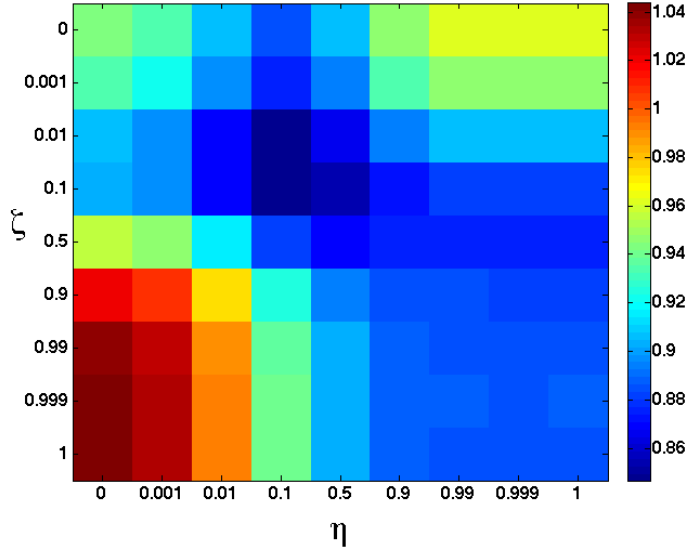}
        \end{center} 
\caption{A heat plot of performance for a range of kernel parameter choices, $\eta$ and $\zeta$, using the MovieLens dataset. }
\label{fig:movielensheat} 
\end{figure}

Figure~\ref{fig:movielensheat} shows the predictive accuracy in RMSE on the MovieLens dataset, obtained by 10-fold cross-validation. The heat plot provides some insight on the relative value, for both movies and users, of the given attribute kernels versus the simple identity kernels. The corners have higher values than some of the values inside the square, showing that the best balance between attribute and Dirac kernels is achieved for $\eta, \zeta \in (0,1)$.

\note{FB: reviewers want more details: e.g., dimension of the feature spaces, which rank constraints, loss function, more details on optimization. We can get away with crappy simulations, but we have to explain them better...\inote{JA: I think this is much better now}
\inote{FB: I agree}}
 
\section{Conclusions}

We have presented a method for solving a generalized matrix completion problem where we have
attributes describing the matrix dimensions. The problem is formalized as the problem of
inferring a linear compact operator between two general Hilbert spaces, which generalizes the
classical finite-dimensional matrix completion problem. We introduced the notion of spectral
regularization for operators, which generalized various spectral penalizations for matrices,
and proved a general representer theorem for this setting. Various approaches, such as standard
low rank matrix completion, are special cases of our method. It is particularly relevant for CF
applications where attributes are available for users and/or objects, and preliminary
experiments confirm the benefits of our method.

An interesting direction of future research is to explore further the multi-task learning
algorithm we obtained with low-rank constraint, and to study the possibility to derive on-line
implementations that may better fit the need for large-scale applications where training data
are continuously increasing. On the theoretical side, a better understanding of the effects of
norm and rank regularizations and their interaction would be of considerable interest.

\appendix
\section{Compact operators on Hilbert spaces}
\label{app:functional}
In this appendix, we recall basic definitions and properties of Hilbert space operators. We refer the interested reader to general books~\citep{Brezis:1980,rkhs} for more details.

Let $\Xcal$ and $\Ycal$ be two Hilbert spaces, with respective inner products denoted by $\inpH{\xb,\xb'}{\Xcal}$ and $\inpH{\yb,\yb'}{\Ycal}$ for $\xb,\xb'\in\Xcal$ and $\yb,\yb'\in\Ycal$. We denote by $\Bcal\br{\Ycal,\Xcal}$ the set of bounded operators from $\Xcal$ to $\Ycal$, i.e., of continuous linear mappings from $\Ycal$ to $\Xcal$. For any two elements $\br{\xb,\yb}$ in $\Xcal\times\Ycal$, we denote by $\xb\otimes\yb$ their \emph{tensor product}, i.e., the linear operator from $\Ycal$ to $\Xcal$ defined by:
\begin{equation}\label{eq:tobject}
\forall \hb \in \Ycal\,,\quad \br{\xb\otimes\yb} \hb = \inpH{\yb,\hb}{\Ycal}\xb\,.
\end{equation}
We denote by $\Bcal_{0}\br{\Ycal,\Xcal}$ the set of \emph{compact} linear operators from $\Ycal$ to $\Xcal$, i.e., the set of linear operators that map the unit ball of $\Ycal$ to a relatively compact set of $\Xcal$. Alternatively, they can also be defined as the limit of finite rank operators.

When $\Xcal$ and $\Ycal$ have finite dimensions, then $\Bcal_{0}\br{\Ycal,\Xcal}$ is simply the set of linear mappings from $\Ycal$ to $\Xcal$, which can be represented by the set of matrices of dimensions $\dim\br{\Xcal}\times\dim\br{\Ycal}$. In that case the tensor product $x\otimes y$ is represented by the matrix $x y^\top$, where $y^\top$ denotes the transpose of $y$. 

For general Hilbert spaces $\Xcal$ and $\Ycal$, any compact linear operator $F \in \Bcal_{0}\br{\Ycal,\Xcal}$ admits a \emph{spectral decomposition}:
\begin{equation}\label{eq:svd}
F = \sum_{i=1}^\infty \sigma_{i} \ub_{i}\otimes \vb_{i}\,.
\end{equation}
Here the the {\em singular values} $(\sigma_{i})_{i\in\NN}$ form a sequence of non-negative real numbers such that $\displaystyle \lim_{i \to \infty} \sigma_i = 0$, and $\br{\ub_{i}}_{i\in\NN}$ and $\br{\vb_{i}}_{i\in\NN}$ form orthonormal families in $\Xcal$ and $\Ycal$, respectively. Although the vectors $\br{\ub_{i}}_{i\in\NN}$ and $\br{\vb_{i}}_{i\in\NN}$ in (\ref{eq:svd}) are not uniquely defined for a given operator $F$, the set of singular values is uniquely defined. By convention we denote by $\sigma_{1}(F), \sigma_{2}(F), \ldots$, the successive singular values of $F$ ranked by decreasing order. The \emph{rank} of $F$ is the number  $\rank(F)\in\NN\cup\cbr{+\infty}$ of strictly positive singular values.

We now describe three subclasses of compact operators of particular relevance in the rest of this paper.

\begin{itemize}
\item The set of operators with finite rank is denoted $\Bcal_{F}\br{\Ycal,\Xcal}$.
\item
The operators $F\in\Bcal_{0}\br{\Ycal,\Xcal}$ that satisfy:
$$
\sum_{i=1}^\infty \sigma_{i}(F)^2 < \infty\,
$$
are called \emph{Hilbert-Schmidt} operators. They form a Hilbert space, denoted $\Bcal_{2}\br{\Ycal,\Xcal}$, with inner product $\inpH{\cdot,\cdot}{\Xcal\otimes\Ycal}$ between basic tensor products given by:
\begin{equation}\label{eq:tensorinp}
\inpH{\xb\otimes\yb , \xb'\otimes\yb'}{\Xcal\otimes\Ycal} = \inpH{\xb,\xb'}{\Xcal}\inpH{\yb,\yb'}{\Ycal}\,.
\end{equation}
In particular, the Hilbert-Schmidt norm of an operator in $\Bcal_{2}\br{\Ycal,\Xcal}$ is given by:
$$
\nm{F}_{2} = \br{\sum_{i=1}^\infty \sigma_{i}(F)^2}^{\frac{1}{2}}\,.
$$
Another useful characterization of Hilbert-Schmidt operators is the following. Each linear operator $F:\Ycal\rightarrow\Xcal$ uniquely defines a bilinear function $f_{H}:\Xcal\times\Ycal\rightarrow\RR$ by
$$
f\br{\xb,\yb} = \inpH{\xb,F \yb}{\Xcal}\,.
$$
The set of functions $f_{F}$ associated to the Hilbert-Schmidt operators forms itself a Hilbert space of functions $\Xcal\times\Ycal\rightarrow\RR$, which is the reproducing kernel Hilbert space of the product kernel defined for $\br{\br{\xb,\yb},\br{\xb',\yb'}}\in\br{\Xcal\times\Ycal}^2$ by
$$
k_{\otimes} \br{\br{\xb,\yb},\br{\xb',\yb'}} = \inpH{\xb,\xb'}{\Xcal}\inpH{\yb,\yb'}{\Ycal}\,.
$$
\item The operators $F\in\Bcal_{0}\br{\Ycal,\Xcal}$ that satisfy:
$$
\sum_{i=1}^\infty \sigma_{i}(F) < \infty\,
$$
are called \emph{trace-class} operators. The set of trace-class operators is denoted $\Bcal_{1}\br{\Ycal,\Xcal}$. The \emph{trace norm} of an operator $F \in \Bcal_{1}\br{\Ycal,\Xcal}$ is given by:
$$
\nm{F}_{1} = \sum_{i=1}^\infty \sigma_{i}(F)\,.
$$
\end{itemize}
Obviously the following ordering exists among these various classes of operators:
$$
\Bcal_{F}\br{\Ycal,\Xcal} \subset \Bcal_{1}\br{\Ycal,\Xcal} \subset \Bcal_{2}\br{\Ycal,\Xcal} \subset \Bcal_{0}\br{\Ycal,\Xcal} \subset \Bcal\br{\Ycal,\Xcal}\,,
$$
and all inclusions are equalities if $\Xcal$ and $\Ycal$ have finite dimensions.

\section{Proof of Theorem \ref{th:representerspectral}}
\label{app:proof}

We start with a general result about the decrease of singular values for compact operators composed with projection:
\begin{lemma}\label{lem:sv}
Let $\Gcal$ and $\Hcal$ be two Hilbert spaces, $H$ a compact linear subspace of $\Hcal$, and $\Pi_{H}$ denote the orthogonal projection onto $H$. Then for any compact operator $F:\Gcal\mapsto\Hcal$ it holds that:
$$
\forall i\geq 1\,,\quad \sigma_{i}(\Pi_{H} F) \leq \sigma_{i}(F)\,.
$$
\end{lemma}
\begin{proof}
We use the classical characterization of the $i$-th singular value:
$$
\sigma_{i}(F) = \max_{V\in\Vcal_{i}(\Gcal)} \min_{\xb\in V , \nm{\xb}_{\Gcal}=1} \nm{F\xb}_{\Hcal}\,,
$$
where $\Vcal_{i}(\Gcal)$ denotes the set of all linear subspaces of $\Gcal$ of dimension $i$. Now, observing that for any $\xb$ we have $\nm{\Pi_{H} F\xb}_{\Hcal} \leq \nm{F\xb}_{\Hcal}$ proves the Lemma.
\end{proof}
Given a training set of patterns $\br{\xb_{i},\yb_{i}}_{i=1,\ldots,N} \in\Xcal\times\Ycal$, remember that we denote by $\Xcal_{N}$ and $\Ycal_{N}$ the linear subspaces of $\Xcal$ and $\Ycal$ spanned by the training patterns $\cbr{\xb_{i}\, , \,i=1,\ldots,N}$ and $\cbr{\yb_{i}\,, \,i=1,\ldots,N}$, respectively. For any operator $F \in \Bcal_{0}\br{\Ycal,\Xcal}$, let us now consider the operator $G = \Pi_{\Xcal_{N}} F  \Pi_{\Ycal_{N}}$. By construction, $F$ and $G$ agree on the training patterns, in the sense that for $i=1,\ldots,N$:
$$
\inpH{\xb_{i},G\yb_{i}}{\Xcal} = \inpH{\xb_{i},\Pi_{\Xcal_{N}} F  \Pi_{\Ycal_{N}}\yb_{i}}{\Xcal} = \inpH{\Pi_{\Xcal_{N}}\xb_{i}, F  \Pi_{\Ycal_{N}}\yb_{i}}{\Xcal} = \inpH{\xb_{i},F\yb_{i}}{\Xcal}\,.
$$
Therefore $F$ and $G$ have the same empirical risk: 
\begin{equation}\label{eq:gf}
R_{N}(F) = R_{N}(G)\,.
\end{equation}
Now, by denoting $F^*$ the adjoint operator, we can use Lemma \ref{lem:sv} and the fact that the singular values of an operator and its adjoint are the same to obtain, for any $i\geq 1$:
\begin{equation*}
\begin{split}
\sigma_{i}(G) &= \sigma_{i}(\Pi_{\Xcal_{N}} F  \Pi_{\Ycal_{N}}) \\
&\leq \sigma_{i}( F  \Pi_{\Ycal_{N}}) \\
&= \sigma_{i}(   \Pi_{\Ycal_{N}} F^*) \\
&\leq \sigma_{i}( F^* ) \\
&= \sigma_{i}( F ).
\end{split}
\end{equation*}
This implies that the spectral penalty term satisfies $\Omega(G)\leq\Omega(F)$. Combined with (\ref{eq:gf}), this shows that if $F$ is a solution to (\ref{eq:optspectral}), then $G = \Pi_{\Xcal_{N}} F  \Pi_{\Ycal_{N}}$ is also a solution. Observing that $G\in\Xcal_{N}\otimes\Ycal_{N}$ concludes the proof of the first part of Theorem \ref{th:representerspectral}, resulting in (\ref{eq:expansion}).

We have now reduced the optimization problem in $\Bcal_{0}\br{\Ycal,\Xcal}$ to a finite-dimensional optimization over the matrix $\alpha$ of size $m_{\Xcal}\times m_{\Ycal}$. Let us now rephrase the optimization problem in this finite-dimensional space.

Let us first consider the spectral penalty term $\Omega(F)$. Given the decomposition (\ref{eq:expansion}), the non-zero singular values of $F$ as an operator are exactly the non-zero singular values of $\alpha$ as a matrix, as soon as $\br{\ub_{1},\ldots,\ub_{m_{\Xcal}}}$ and $\br{\vb_{1},\ldots,\vb_{m_{\Ycal}}}$ form \emph{orthonormal} bases of $\Xcal_{N}$ and $\Ycal_{N}$, respectively. In order to be able to express the empirical risk $R_{N}(F)$ we must however consider a decomposition of $F$ over the training patterns, as:
\begin{equation}\label{eq:expansiongamma}
F = \sum_{i=1}^N \sum_{j=1}^N \gamma_{ij} \xb_{i}\otimes\yb_{j}\,.
\end{equation}
In order to express the singular values from this expression let us introduce the \emph{Gram matrices} $K$ and $G$ of the training patterns, i.e., the $N\times N$ matrices defined for $i,j=1,\ldots,N$ by:
$$
K_{ij} = \inpH{\xb_{i},\xb_{j}}{\Xcal}\,,\quad G_{ij} = \inpH{\yb_{i},\yb_{j}}{\Ycal}\,.
$$
We note that by definition the ranks of $K$ and $G$ are respectively $m_{\Xcal}$ and $m_{\Ycal}$. Let us now factorize these two matrices as $K=XX^\top$ and $G=YY^\top$,  where $X\in\RR^{N\times m_{\Xcal}}$ and $Y\in\RR^{N\times m_{\Ycal}}$ are any square roots, e.g., obtained by kernel PCA or Cholesky decomposition~\citep{fine01efficient,csi}. The matrices $X$ and $Y$ provide a representation of the pattern in two orthonormal bases which we denote by $\br{\ub_{1},\ldots,\ub_{m_{\Xcal}}}$ and $\br{\vb_{1},\ldots,\vb_{m_{\Ycal}}}$. In particular we have, for any $i,j \in 1,\ldots,N$:
$$
\xb_{i}\otimes\yb_{j} = \sum_{l=1}^{m_{\Xcal}} \sum_{m=1}^{m_{\Ycal}} X_{il}Y_{jm} \ub_{l}\otimes \vb_{m}\,,
$$
from which we deduce:
$$
F = \sum_{l=1}^{m_{\Xcal}} \sum_{m=1}^{m_{\Ycal}} \br{\sum_{i=1}^N \sum_{j=1}^N X_{il} Y_{jm} \gamma_{ij}} \ub_{l}\otimes \vb_{m}\,.
$$
Comparing this expression to (\ref{eq:expansion}) we deduce that:
$$
\alpha = X^\top \gamma Y\,.
$$
The empirical error $R_{N}(F)$ is a function of $f\br{\xb_{l},\yb_{l}}$ for $l=1,\ldots,N$. From (\ref{eq:expansiongamma}), we see that:
$$
f\br{\xb_{l},\yb_{l}} = \sum_{i=1}^N \sum_{j=1}^N \gamma_{ij} K_{il} G_{lj}\,,
$$
and therefore the vector of predictions $F_{N} = \br{f\br{\xb_{l},\yb_{l}}}_{l=1,\ldots,N} \in \RR^N$ can be rewritten as:
$$
F_{N} = \diag(K\gamma G) = \diag\br{X \alpha Y^\top}\,.
$$
We can now replace the empirical risk $R_{N}(F_{N})$ by $R_{N}\br{\diag\br{X \alpha Y^\top}}$ and the penalty $\Omega(F)$ by $\Omega(\alpha)$ to deduce the optimization problem (\ref{eq:optalpha}) from (\ref{eq:optspectral}), which concludes the proof of Theorem \ref{th:representerspectral}.

\section{Proof of Proposition~\ref{prop:conv_lowrank}}
\label{app:lowrank}

Since the function has compact level sets, we may assume that we are restricted to an open bounded subset of $\rb^{ p \times q}$ where the second and first derivatives are uniformly bounded. We let denote $C>0$ a common upper bound of all derivatives. The gradient of the function $H$ is equal to
$ \nabla H = { \nabla G^\top \,  U \choose \nabla G\ \,  V }$, while the Hessian of $H$ is the following quadratic form:
$$
\nabla^2 H[ (dU,dV),(dU,dV) ]
= 2 \tr dV^\top \nabla G dU
+ \nabla^2 G[ U dV^\top + dU V^\top, U dV^\top + dU V^\top].
$$
Without loss of generality, we may assume that the last columns of $U$ and $V$ are equal to zero (this can be done by rotation of $U$ or $V$).
The zero gradient assumption implies that
$ \nabla G^\top  {U}=0$ and $ \nabla G  {V}=0$. While if we take
$dU$ and $dV$ with the first $m-1$ columns equal to zero, and last columns equal
to arbitrary $u$ and $v$, then the second term in the Hessian is equal to zero. The positivity of the first term implies that for all $u$ and $v$, 
$  v^\top \nabla Gu \geqslant 0$, i.e., the gradient of $G$ at $N=UV^\top$ is equal to zero, and thus we get a stationary point and thus a global minimum of $G$.

\bibliography{lowrank_jmlr}

\begin{thebibliography}{28}
\providecommand{\natexlab}[1]{#1}
\providecommand{\url}[1]{\texttt{#1}}
\expandafter\ifx\csname urlstyle\endcsname\relax
  \providecommand{\doi}[1]{doi: #1}\else
  \providecommand{\doi}{doi: \begingroup \urlstyle{rm}\Url}\fi

\bibitem[Abernethy et~al.(2006)Abernethy, Bach, Evgeniou, and Vert]{lowrank}
J.~Abernethy, F.~Bach, T.~Evgeniou, and J.-P. Vert.
\newblock Low-rank matrix factorization with attributes.
\newblock Technical Report N24/06/MM, Ecole des Mines de Paris, 2006.

\bibitem[Amit et~al.(2007)Amit, Fink, Srebro, and Ullman]{srebro-mc}
Y.~Amit, M.~Fink, N.~Srebro, and S.~Ullman.
\newblock Uncovering shared structures in multiclass classification.
\newblock In \emph{Proceedings of the 24th international conference on Machine
  learning}, pages 17--24, New York, NY, USA, 2007. ACM.

\bibitem[Argyriou et~al.(2008)Argyriou, Evgeniou, and
  Pontil]{Argyriou2008Convex}
A.~Argyriou, T.~Evgeniou, and M.~Pontil.
\newblock Convex multi-task feature learning.
\newblock \emph{Machine Learning}, 2008.
\newblock To appear.

\bibitem[Aronszajn(1950)]{Aronszajn1950Theory}
N.~Aronszajn.
\newblock Theory of reproducing kernels.
\newblock \emph{Trans. {A}m. {M}ath. {S}oc.}, 68:\penalty0 337~--~404, 1950.

\bibitem[Bach(2008)]{bach-tracenorm}
F.~R. Bach.
\newblock Consistency of trace norm minimization.
\newblock \emph{J. Mach. Learn. Res.}, 9:\penalty0 1019--1048, 2008.

\bibitem[Bach and Jordan(2005)]{csi}
F.~R. Bach and M.~I. Jordan.
\newblock Predictive low-rank decomposition for kernel methods.
\newblock In \emph{ICML '05: Proceedings of the 22nd international conference
  on Machine learning}, pages 33--40, New York, NY, USA, 2005. ACM.

\bibitem[Bach et~al.(2004)Bach, Lanckriet, and Jordan]{BachLanc2004skm}
F.~R. Bach, G.~R.~G. Lanckriet, and M.~I. Jordan.
\newblock Multiple kernel learning, conic duality, and the {SMO} algorithm.
\newblock In \emph{ICML '04: Proceedings of the twenty-first international
  conference on Machine learning}, page~6, New York, NY, USA, 2004. ACM.

\bibitem[Bach et~al.(2005)Bach, Thibaux, and Jordan]{bach_thibaux}
F.~R. Bach, R.~Thibaux, and M.~I. Jordan.
\newblock Computing regularization paths for learning multiple kernels.
\newblock In Lawrence~K. Saul, Yair Weiss, and {L\'{e}on} Bottou, editors,
  \emph{Advances in Neural Information Processing Systems 17}, pages 73--80,
  Cambridge, MA, 2005. MIT Press.

\bibitem[Berlinet and Thomas-Agnan(2003)]{rkhs}
A.~Berlinet and C.~Thomas-Agnan.
\newblock \emph{Reproducing Kernel {H}ilbert Spaces in Probability and
  Statistics}.
\newblock Kluwer Academic Publishers, 2003.

\bibitem[Boyd and Vandenberghe(2003)]{Boyd2003Convex}
S.~Boyd and L.~Vandenberghe.
\newblock \emph{Convex Optimization}.
\newblock Cambridge Univ. Press, 2003.

\bibitem[Breese et~al.(1998)Breese, Heckerman, and Kadie]{CF1}
J.~S. Breese, D.~Heckerman, and C.~Kadie.
\newblock Empirical analysis of predictive algorithms for collaborative
  filtering.
\newblock In \emph{14th Conference on Uncertainty in Artificial Intelligence},
  pages 43--52, Madison, W.I., 1998. Morgan Kaufman.

\bibitem[Brezis(1980)]{Brezis:1980}
H.~Brezis.
\newblock \emph{Analyse Fonctionnelle}.
\newblock Masson, 1980.

\bibitem[Burer and Choi(2006)]{burer2}
S.~A. Burer and C.~Choi.
\newblock Computational enhancements in low-rank semidefinite programming.
\newblock \emph{Optimization Methods and Software}, 21:\penalty0 493--512,
  2006.

\bibitem[Burer and Monteiro(2005)]{burer1}
S.~A. Burer and R.~D.~C. Monteiro.
\newblock Local minima and convergence in low-rank semidefinite programming.
\newblock \emph{Mathematical Programming}, 103:\penalty0 427--444, 2005.

\bibitem[Evgeniou et~al.(2005)Evgeniou, Micchelli, and
  Pontil]{Evgeniou2005Learning}
T.~Evgeniou, C.~Micchelli, and M.~Pontil.
\newblock Learning multiple tasks with kernel methods.
\newblock \emph{J. Mach. Learn. Res.}, 6:\penalty0 615--637, 2005.

\bibitem[Fazel et~al.(2001)Fazel, Hindi, and Boyd]{Boyd2001Rank}
M.~Fazel, H.~Hindi, and S.~Boyd.
\newblock A rank minimization heuristic with application to minimum order
  system approximation.
\newblock In \emph{Proceedings of the 2001 American Control Conference},
  volume~6, pages 4734--4739, 2001.

\bibitem[Fine and Scheinberg(2001)]{fine01efficient}
S.~Fine and K.~Scheinberg.
\newblock Efficient {SVM} training using low-rank kernel representations.
\newblock \emph{J. Mach. Learn. Res.}, 2:\penalty0 243--264, 2001.

\bibitem[Golub and Loan(1996)]{golub83matrix}
G.~H. Golub and C.~F.~Van Loan.
\newblock \emph{Matrix Computations}.
\newblock Johns Hopkins University Press, 1996.

\bibitem[Heckerman et~al.(2000)Heckerman, Chickering, Meek, Rounthwaite, and
  Kadie]{CF2}
D.~Heckerman, D.~M. Chickering, C.~Meek, R.~Rounthwaite, and C.~Kadie.
\newblock Dependency networks for inference, collaborative filtering, and data
  visualization.
\newblock \emph{J. Mach. Learn. Res.}, 1:\penalty0 49--75, 2000.

\bibitem[Jacob and Vert(2008)]{Jacob2006Epitope}
L.~Jacob and J.-P. Vert.
\newblock Efficient peptide-{MHC}-{I} binding prediction for alleles with few
  known binders.
\newblock \emph{Bioinformatics}, 24\penalty0 (3):\penalty0 358--366, Feb 2008.

\bibitem[Lanckriet et~al.(2004)Lanckriet, Cristianini, Ghaoui, Bartlett, and
  Jordan]{gert}
G.~R.~G. Lanckriet, N.~Cristianini, L.~El Ghaoui, P.~Bartlett, and M.~I.
  Jordan.
\newblock Learning the kernel matrix with semidefinite programming.
\newblock \emph{Journal of Machine Learning Research}, 5:\penalty0 27--72,
  2004.

\bibitem[Lewis and Sendov(2002)]{lewis02twice}
A.~S. Lewis and H.~S. Sendov.
\newblock Twice differentiable spectral functions.
\newblock \emph{SIAM J. Mat. Anal. App.}, 23\penalty0 (2):\penalty0 368--386,
  2002.

\bibitem[Rennie and Srebro(2005)]{Rennie2005Fast}
J.~D.~M. Rennie and N.~Srebro.
\newblock Fast maximum margin matrix factorization for collaborative
  prediction.
\newblock In \emph{Proceedings of the 22nd international conference on Machine
  learning}, pages 713--719, New York, NY, USA, 2005. ACM Press.

\bibitem[Salakhutdinov et~al.(2007)Salakhutdinov, Mnih, and Hinton]{CF3}
R.~Salakhutdinov, A.~Mnih, and G.~Hinton.
\newblock Restricted boltzmann machines for collaborative filtering.
\newblock In \emph{ICML '07: Proceedings of the 24th international conference
  on Machine learning}, pages 791--798, New York, NY, USA, 2007. ACM.

\bibitem[Sch{\"o}lkopf et~al.(2001)Sch{\"o}lkopf, Herbrich, and
  Smola]{representer}
B.~Sch{\"o}lkopf, R.~Herbrich, and A.~J. Smola.
\newblock A generalized representer theorem.
\newblock In \emph{Proceedings of the 14th Annual Conference on Computational
  Learning Theory}, volume 2011 of \emph{Lecture Notes in Computer Science},
  pages 416--426, Berlin / Heidelberg, 2001. Springer.

\bibitem[Shawe-Taylor and Cristianini(2004)]{Cristianini2004}
J.~Shawe-Taylor and N.~Cristianini.
\newblock \emph{Kernel Methods for Pattern Analysis}.
\newblock Cambridge University Press, 2004.

\bibitem[Srebro and Jaakkola(2003)]{Srebro2003Weighted}
N.~Srebro and T.~Jaakkola.
\newblock Weighted low-rank approximations.
\newblock In T.~Fawcett and N.~Mishra, editors, \emph{Proceedings of the
  Twentieth International Conference on Machine Learning}, pages 720--727. AAAI
  Press, 2003.

\bibitem[Srebro et~al.(2005)Srebro, Rennie, and Jaakkola]{Srebro2005Maximum}
N.~Srebro, J.~D.~M. Rennie, and T.~S. Jaakkola.
\newblock Maximum-margin matrix factorization.
\newblock In L.~K. Saul, Y.~Weiss, and L.~Bottou, editors, \emph{Adv. Neural.
  Inform. Process Syst. 17}, pages 1329--1336, Cambridge, MA, 2005. MIT Press.

\end{thebibliography}

\end{document}